\documentclass[11pt,a4paper]{article}
\PassOptionsToPackage{hyphens}{url}\usepackage[hyperref]{naaclhlt2019}
\usepackage{times}
\usepackage{latexsym}
\usepackage{todonotes}
\usepackage{url}
\usepackage{booktabs} 

\newcommand{\Ni}{({\em i})~}
\newcommand{\Nii}{({\em ii})~}

\newcommand{\good}{\textsc{Good}\,}
\newcommand{\bad}{\textsc{Bad}\,}
\newcommand{\potential}{\textsc{Potentially Useful}\,}

\aclfinalcopy 

\title{SemEval-2019 Task 8:\\  Fact Checking in Community Question Answering Forums}

\author{Tsvetomila Mihaylova,\textsuperscript{1}
Georgi Karadjov,\textsuperscript{2}
Pepa Atanasova,\textsuperscript{3}\\
{\bf Ramy Baly,\textsuperscript{4}
Mitra Mohtarami,\textsuperscript{4}
Preslav Nakov\textsuperscript{5}} \\
\textsuperscript{1} Instituto de Telecomunica\c{c}\~oes, Lisbon, Portugal,
\textsuperscript{2} SiteGround Hosting EOOD, Bulgaria \\
\textsuperscript{3} University of Copenhagen, Denmark\\
\textsuperscript{4} MIT Computer Science and Artificial Intelligence Laboratory, Cambridge, MA\\
\textsuperscript{5} Qatar Computing Research Institute, HBKU\\
  {\tt \{tsvetomila.mihaylova, georgi.m.karadjov\}@gmail.com}, \\ {\tt pepa@di.ku.dk},
  {\tt \{baly,mitram\}@mit.edu},
  {\tt pnakov@qf.org.qa}
  }

\date{}

\begin{document}

\maketitle
\begin{abstract}
We present SemEval-2019 Task 8 on \emph{Fact Checking in Community Question Answering Forums}, which features two subtasks.
Subtask A is about deciding whether a question asks for factual information vs. an opinion/advice vs. just socializing.
Subtask B asks to predict whether an answer to a factual question is true, false or not a proper answer.
We received 17 official submissions for subtask A and 11 official submissions for Subtask B. For subtask A, all systems improved over the majority class baseline. For Subtask B, all systems were below a majority class baseline, but several systems were very close to it.
The leaderboard and the data from the competition can be found at \url{http://competitions.codalab.org/competitions/20022}.
\end{abstract}

\section{Overview}

The current coverage of the political landscape in both the press and in social media has led to an unprecedented situation. Like never before, a statement in an interview, a press release, a blog note, or a tweet can spread almost instantaneously. 
The speed of proliferation leaves little time for double-checking claims against the facts, which has proven critical in politics, e.g., during the 2016 presidential campaign in the USA, which was dominated by fake news in social media and by false claims.

Investigative journalists and volunteers have been working hard to get to the root of a claim and to present solid evidence in favor or against it. Manual fact-checking is very time-consuming, and thus automatic methods have been proposed to speed-up the process, e.g.,~there has been work on checking the factuality/credibility of a claim, of a news article, or of an information source~\cite{ba2016vera, zubiaga2016analysing, ma2016detecting, castillo2011information,D18-1389}.

\noindent The process starts when a document is made public. First, an intrinsic analysis is carried out in which check-worthy text fragments are identified.
Then, other documents that might support or rebut a claim in the document are retrieved from various sources. Finally, by comparing a claim against the retrieved evidence, a system can determine whether the claim is likely true or likely false (or unsure, if no strong enough evidence either way could be found). For instance, \citet{10.1371journal.pone.0128193} do this using a knowledge graph derived from Wikipedia. The outcome could then be presented to a human expert for final judgement.\footnote{As of present, fully automatic methods for fact checking still lag behind in terms of quality, and thus also of credibility in the eyes of the users, compared to what high-quality manual checking by reputable sources can achieve, which means that a final double-checking by a human expert is needed.}

For our two subtasks, we explore factuality in the context of Community Question Answering (cQA) forums. Forums such as StackOverflow, Yahoo! Answers, and Quora are very popular these days, as they represent effective means for communities around particular topics to share information.
However, the information shared by the users is not always correct or accurate. There are multiple factors explaining the presence of incorrect answers in cQA forums, e.g., misunderstanding of the question, ignorance or maliciousness of the responder. Also, as a result of our dynamic world, the truth is time-sensitive: something that was true yesterday may be false today. Moreover, forums are often barely moderated and thus lack systematic quality control.

Here we focus on checking the factuality of questions and answers in cQA forums. This aspect was ignored in recent cQA tasks~\cite{Ishikawa:10,nakov-EtAl:2015:SemEval,nakov-EtAl:2016:SemEval,SemEval-2017:task3}, where an answer is considered \good\ if it addresses the question, irrespective of its veracity, accuracy, etc.

\noindent Figure~\ref{fig:example1} presents an excerpt of an example from the Qatar Living Forum, with one question and three answers selected from a longer thread.
According to SemEval-2016 Task 3~\cite{nakov-EtAl:2016:SemEval}, all three answers would be considered \good\ since they are formally answering the question. Nevertheless, $a_1$ contains false information, while $a_2$ and $a_3$ are correct, as can be established from an official government website.\footnote{\url{http://portal.moi.gov.qa/wps/portal/MOIInternet/departmentcommittees/visasentrypermeits/}}

\begin{figure}[t]
\small
\begin{itemize}
\setlength\itemsep{0em}
\item[$Q$:] HI ;; IF WIFE IS UNDER HER HUSBAND'S SPONSORSHIP AND IS WILLING TO COME QATAR ON VISIT; HOW LONG SHE CAN STAY AFTER EXTENDING THE VISA EVERY MONTH? I HAVE HEARD ITS NOT POSSIBLE TO EXTEND VISIT VISA MORE THAN 6 MONTHS? CAN U PLEASE ANSWER ME.. THANKZZZ...
\item[$a_1$:] Maximum period is 9 Months....
\item[$a_2$:] 6 months maximum
\item[$a_3$:] This has been answered in QL so many times. Please do search for information regarding this. BTW answer is 6 months.
\end{itemize}
\vspace{-6pt}
\caption{\label{fig:example}Example from the \emph{Qatar Living} forum.}
\label{fig:example1}
\end{figure}

Checking the veracity of answers in a cQA forum is a hard problem, which requires putting together aspects of language understanding, modelling the context, integrating several information sources, uisng world knowledge and complex inference, among others.
Moreover, high-quality automatic fact-checking would offer better experience to users of cQA systems,
e.g., the user could be presented with veracity scores, where low scores would warn the user
not to completely trust the answer or to double-check it.

\section{Related Work}

Fact-checking of answers was not studied before in the context of community Question Answering, apart from our own recent work \cite{AAAI2018:factchecking}.
Yet, in the context of cQA and general QA, there has been work on \emph{credibility} assessment, which has been modelled primarily at the feature level, with the goal of improving \good\ answer identification. A notable exception are \cite{RANLP2017:credibility:trolls,InternetResearchJournal:2018}, where credibility was a task on its own right.
However, \emph{credibility} is different from \emph{veracity} (our focus here) as it is a subjective perception about whether a statement is credible, rather than actually truthful.

\noindent \newcite{Jurczyk:2007:DAQ:1321440.1321575} modelled author authority using link analysis.
\newcite{Agichtein:2008:FHC:1341531.1341557} looked for high-quality answers using PageRank and HITS, in addition to intrinsic content quality, e.g., punctuation and typos, syntactic and semantic complexity, and grammaticality.

\newcite{lita2005qualitative} studied three qualitative dimensions for answers:
 source credibility (e.g.,~does the document come from a government website),
 sentiment analysis, and contradiction compared to other answers.
\newcite{Su-EtAl:2010:PACLIC2010} looked for verbs and adjectives that cast doubt.
\newcite{banerjee-han:2009:NAACLHLT09-Short} used
language modelling to validate the reliability of an answer's source.
\newcite{Jeon:2006:FPQ:1148170.1148212} focused on non-textual features such as click counts,
answer activity level, and copy counts.
\newcite{Pelleg:2016:CEI:2818048.2820022} curated social media content using syntactic, semantic, and social signals.
Unlike this research, we \Ni target factuality rather than credibility, \Nii address it as a task in its own right, and on a specialised dataset.

Information credibility was also studied in social computing. \newcite{castillo2011information} modeledd user reputation.
\newcite{Canini:2011} analyzed the interaction of content and social network structure.
\newcite{Morris:2012:TBU:2145204.2145274} studied how Twitter users judge truthfulness.
\newcite{lukasik-cohn-bontcheva:2015:ACL-IJCNLP} used temporal patterns to detect rumors,
and \newcite{zubiaga2016analysing} focused on conversations.

Other authors have been querying the Web to gather support for accepting or refuting a claim \cite{popat2016credibility,RANLP2017:factchecking:external}.
In social media, there has been research targeting the user, e.g., finding malicious users \cite{ACL2016:trolls,AAAI2018:factchecking,InternetResearchJournal:2018}, \emph{sockpuppets} \cite{Maity:2017:DSS:3022198.3026360}, \emph{Internet water army} \cite{Chen:2013:BIW:2492517.2492637}, and \emph{seminar users} \cite{SeminarUsers2017}.

Finally, there has been work on credibility, trust, and expertise in
news communities \cite{mukherjee2015leveraging}.
\citet{Dong:2015:KTE:2777598.2777603} proposed that a trustworthy source is one that contains very few false claims.
Recent work has also focused on evaluating the factuality of reporting of entire news outlets \cite{D18-1389,source:multitask:NAACL:2019}.\footnote{Knowing the reliability of a medium is important when fact-checking a claim \cite{Popat:2017:TLE:3041021.3055133,DBLP:conf/aaai/NguyenKLW18} and when solving article-level tasks such as ``fake news'' and click-bait detection \cite{Hardalov2016,RANLP2017:clickbait,Pan:KG:2018,prezrosas-EtAl:2018:C18-1}.}
However, none of this work was about QA or cQA.

\section{Subtasks and Data Description}

SemEval-2019 Task 8 has two subtasks:

\begin{itemize}
\item {\bf Subtask~A:} Given a question from a cQA forum, predict whether this question asks for factual information vs. opinion/advice vs. just socializing.
\item {\bf Subtask~B:} Given a factual question from a cQA forum, together with its answer thread, predict whether each answer provides true vs. false vs. non-factual information as a response to the question.
\end{itemize}

\subsection{Data and Resources}

We retrieved the data from the Qatar Living web forum\footnote{\url{http://www.qatarliving.com}}. We then cleaned it and we annotated it with the labels described in Sections \ref{section:data_subtask_a} and \ref{section:data_subtask_b}.

For subtask A, we annotated the questions using Amazon Mechanical Turk\footnote{\url{http://www.mturk.com/}}. To ensure high quality of the annotation, we went through all annotations and manually double-checked them.

For subtask B, we did not use an external annotation service, but instead we annotated all the data ourselves. Each answer was processed by three independent annotators, and we made sure we had proof for the label from reliable sources on the Web. Then, the annotations were consolidated after a discussion until agreement was achieved for each example.

All data is freely available under a Creative Commons Attribution 3.0 Unported (CC BY 3.0) license, and is accessible on the competition's website\footnote{\url{http://competitions.codalab.org/competitions/20022}}.

In addition to the provided annotated data, we also allowed the participants to use unlabelled data from the Qatar Living forum footnote{\url{http://alt.qcri.org/semeval2016/task3/data/uploads/QL-unannotated-data-subtaskA.xml.zip}}, as well as additional external resources, which they had to mention explicitly in their submissions.

Note that the class distribution in the training, development and test sets differs, especially for Subtask B.
The reason for this is the way the data was prepared. The different datasets (training, development and test) were prepared on stages, because of the very time-consuming data annotation process.

\noindent For each dataset annotation stage, we had to choose between releasing all the available annotated data or aim at releasing sets with similar label distribution. At the end, we decided to release the available data, although we were aware that this would result in releasing sets with different distribution and, in some cases, unbalanced categories.

\subsection{Training Data for Subtask A}\label{section:data_subtask_a}
To create the dataset for the task, we chose to augment a pre-existing dataset for cQA with factuality annotations; this allowed us to stress the difference between ({\it a})~distinguishing a good vs. a bad answer, and ({\it b})~distinguishing a factually-true vs. a factually-false one.
In particular, we added annotations for factuality to the CQA-QL-2016 dataset from SemEval-2016 Task~3 on Community Question Answering~\cite{nakov-EtAl:2016:SemEval}.

In CQA-QL-2016, the data is organized in question--answer threads (from the Qatar Living forum).
Each question has a subject, a body, and meta information: question ID, date and time of posting, user name and ID, and category (e.g.,~{\it Computers and Internet} and {\it Moving to Qatar}).

We analyzed the forum questions and we defined three categories, related to their factuality. We then annotated the questions using Amazon Mechanical Turk. The three factuality categories are as follows:

\begin{itemize}

\item[$\ast$]
\textsc{Factual}:
The question asks for factual information, which can be answered by checking various information sources, and it is not ambiguous
(e.g., ``{\it What is Ooredoo customer service number?}'').

\item[$\ast$] \textsc{Opinion}: The question asks for an opinion or an advice, not for a fact.
(e.g., ``{\it Can anyone recommend a good Vet in Doha?}'')

\item[$\ast$] \textsc{Socializing}:
Not a real question, but rather socializing/chatting.
This can also mean expressing an opinion or sharing some information, without really asking anything of general interest.
(e.g., ``{\it What was your first car?}'')

\end{itemize}

Table~\ref{table:question-labels-distribution} shows the distribution of the labels for the question labels in the training, in the development and in the testing datasets. Overall, there are 1118, 239 and 953 questions annotated with the above-described labels.

\begin{table}[t]
\small
\centering
\begin{tabular}{p{1.5in}rrr}
\bf Label & \bf Train & \bf Dev & \bf Test \\
\hline \\
\textsc{Factual} & 311 & 62 & 299 \\
\textsc{Opinion} & 563 & 126 & 167 \\
\textsc{Socializing} & 244 & 51 & 487 \\
\hline \\
 \bf \textsc{Total} & \bf 1118 & \bf 239 & \bf 953 \\

\end{tabular}
\caption{\textbf{Subtask A:} Distribution of the factuality labels for the questions.}
\label{table:question-labels-distribution}
\end{table}

\subsection{Training Data for Subtask B}\label{section:data_subtask_b}

For subtask~B, we annotated for veracity the answers to the questions with a
\textsc{Factual} label for subtask A.
Note that in CQA-QL-2016, each answer has a subject, a body, meta information (answer ID, user name, and ID), the question that it answers, and a judgement about how well it answers the question of its thread (\good, \bad\ or \potential).

We annotated the \good\ answers for factuality based on the assumption that a \good\ answer means it provides factual information, whether it is true or false. All \bad\ and \potential answers are automatically considered as \textsc{Non-Factual}.
The factuality labels are described as follows:

\begin{itemize}

\item[$\ast$] \textsc{Factual -- True}:
The answer is True and can be proven with an external resource.
({\bf Q:}~{\it ``I wanted to know if there were any specific shots and vaccinations I should get before coming over [to Doha].''};
{\bf A:}~{\it ``Yes there are; though it varies depending on which country you come from. In the UK; the doctor has a list of all countries and the vaccinations needed for each.''}).\footnote{The answer is factually true and this can be seen at \url{http://wwwnc.cdc.gov/travel/destinations/traveler/none/qatar}}

\item[$\ast$] \textsc{Factual -- False}:
The answer gives a factual response, but it is False and this can be proven using an external resource.
({\bf Q:}~{\it ``Can I bring my pitbulls to Qatar?''};
{\bf A:}~{\it ``Yes you can bring it but be careful this kind of dog is very dangerous.''}).\footnote{The answer is incorrect since pitbulls are included in the list of breeds banned in Qatar. See \url{http://canvethospital.com/pet-relocation/banned-dog-breed-list-qatar-2015/}}

\item[$\ast$] \textsc{Factual -- Partially True}:
The answer contains more than one claim, and only some of these claims could be manually verified.
({\bf Q:}~{\it ``I will be relocating from the UK to Qatar [...] is there a league or TT clubs / nights in Doha?''};
{\bf A:}~{\it ``Visit Qatar Bowling Center during thursday and friday and you'll find people playing TT there.''}).\footnote{The place mentioned in the answer has table tennis, but we do not know on which days. See \url{http://www.qatarbowlingfederation.com/bowling-center/}}

\item[$\ast$] \textsc{Factual -- Conditionally True}:
The answer is True in some cases, and False in others, depending on some conditions that the answer does not mention.
({\bf Q:}~{\it ``My wife does not have NOC from Qatar Airways; but we are married now so can i bring her legally on my family visa as her husband?''};
{\bf A:}~{\it ``Yes you can.''}).\footnote{This answer can be true, but this depends upon some conditions. See \url{http://www.onlineqatar.com/info/dependent-family-visa.aspx}}

\item[$\ast$] \textsc{Factual - Responder Unsure}:
The person giving the answer is not sure about the veracity of his/her statement.
(e.g., ``{\it Possible only if government employed. That's what I heard.}'')

\item[$\ast$] \textsc{Non-Factual}:  When the answer does not provide factual information to the question;
it can be an opinion or an advice that cannot be verified.
(e.g., ``{\it Its better to buy a new one.}'').
\end{itemize}

\begin{table}[t]
\small
\centering
\begin{tabular}{p{1.5in}rrr}
\bf Label & \bf Train & \bf Dev & \bf Test \\
\hline \\
\textsc{True}  & 166 & 29 & 34 \\
\textsc{False} & 135 & 31 & 45 \\
\textsc{NonFactual} & 194 & 52 & 231 \\
\hline \\
 \bf \textsc{Total} & \bf 495 & \bf 112 & \bf 310 \\

\end{tabular}
\caption{\textbf{Subtask B:} Distribution of the factuality labels for the answers.}
\label{table:answer-labels-distribution}
\end{table}

We further discarded answers whose factuality was very time-sensitive and it makes no sense to check whether the statements are true or false (e.g., ``\emph{It is Friday tomorrow.}'', ``\emph{It was raining last week.}'').

Moreover, many answers are arguably somewhat time-sensitive, e.g., ``\emph{There is an IKEA in Doha.}'' is true only after IKEA opened, but not before that. In such cases, we just used the present situation as a point of reference. We further discarded the answers for which the annotators could not find any information.

\noindent Ultimately, we consolidated the above fine-grained labels into the following coarse-grained labels, which we used for subtask B:

\begin{itemize}
    \item[$\ast$] \textsc{Factual -- True}: Contains answers with proven true, non-contradictory statements. This includes the answers with the label \textsc{Factual -- True} from above. This label is used for answers one can trust as a true statement.
    \item[$\ast$] \textsc{Factual -- False}: Contains answers with statements that are proven to be false or not completely true. This includes answers with the following fine-grained factuality labels: \textsc{Factual -- False}, \textsc{Factual -- Partially False}, \textsc{Factual -- Conditionally True}, \textsc{Factual -- Responder Unsure}. We also use this label for answers that contain a statement for which the person giving the answer expresses uncertainty in the claim.
    \item[$\ast$] \textsc{Non-Factual}: These are either non-factual statements or statements that could be factual, but no information about them could be found, i.e.,~we could find no way to check whether the statement was true or false. This category also includes some statements that have been incorrectly annotated as a \good answer. It also includes the very time-sensitive statements described before, such as ''\emph{It is Friday tomorrow?}''. The \bad and the \potential answers from CQA-QL-2016 also fall in this category.
\end{itemize}

As we have mentioned above, we have annotated the answers to the \textsc{Factual} questions selected from the Qatar Living forum.
We targeted very high quality annotation, and thus we did not use crowd-sourcing, as a pilot experiment has shown that the task was very difficult and that it was not possible to guarantee that Turkers would do all the necessary verification and gather evidence from trusted sources.
Instead, all examples were first annotated independently by three of us, and then, we carefully discussed {\it each example} to come up with a final label.
The distribution of the labels on the training, on the development, and on the testind dataset are shown in Table~\ref{table:answer-labels-distribution}\footnote{Although not very big, our dataset is larger than datasets used for similar problems, e.g.,~\newcite{Ma:2015:DRU} experimented with 226 rumors for rumor detection, and~\newcite{popat2016credibility} used 100 Wiki hoaxes for credibility assessment of textual claims.}.

\subsection{Evaluation}

Both subtasks are three-way classification problems.
In subtask A, the questions were to be classified as \textsc{Factual}, \textsc{Opinion}, or \textsc{Socializing}.
Similarly, in subtask B there were also three target categories for the answers: \textsc{Factual - True}, \textsc{Factual - False}, and \textsc{Non-Factual}.

We further scored the submissions based on Accuracy, macro-F1, and average recall (AvgRec).\footnote{Average recall has some attractive properties and has been used in previous SemEval tasks, e.g.,~\cite{nakov-etal-2016-semeval,rosenthal-etal-2017-semeval}.} For subtask B, we also report mean average precision (MAP), where the \textsc{Factual - True} instances were considered to be positive, and the remaining ones were negative.
The official evaluation measure for both subtasks was Accuracy.


\section{Participants and Results}

We received 17 official submissions for Subtask A and 11 official submissions for Subtask B. Below we report the evaluation results.

Table~\ref{table:results_subtask_a} presents the results for subtask A on question classification. The results are based the official submissions in the evaluation phase.
In this subtask, all of the submitted systems managed to improve over the majority class baseline, and several teams achieved similarly good results. Whenever a number of teams achieve the same result with respect to the main evaluation measure, i.e.,~Accuracy, we rank them according to the F1 score, and then by AvgRec if a tie still appears.

\begin{table*}[!ht]
\small
\centering
\begin{tabular}{lp{50mm}cll}
\\
\bf Team ID & \bf Affiliation & \bf Accuracy & \bf F1 & \bf AvgRec \\
\midrule
Fermi \cite{SemEval2019-task8-fermi} & IIIT Hyderabad, Microsoft, Teradata & \bf 0.840 & 0.718$_{2}$ & 0.735$_{3}$ \\
TMLab \cite{SemEval2019-task8-tmlab} &  Samsung R\&D Institute, Warsaw, Poland & \bf 0.834 & 0.725$_{1}$ & 0.764$_{1}$ \\
SolomonLab \cite{SemEval2019-task8-solomonlab}  & Samsung R\&D Institute India, Bangalore & \bf 0.831 & 0.709$_{4}$ & 0.728$_{4}$ \\
ColumbiaNLP  \cite{SemEval2019-task8-columbianlp} & Columbia University, Department Of Computer Science and Data Science Institute & \bf 0.828 & 0.645$_{7}$ & 0.662$_{9}$ \\
DOMLIN \cite{SemEval2019-task8-domlin} & Deutsches Forschungszentrum f\"ur K\"unstliche Intelligenz (DFKI), Saarbrucken, Germany & \bf 0.823 & 0.710$_{3}$ & 0.755$_{2}$ \\
BLCU\_NLP \cite{SemEval2019-task8-blcunlp} & Beijing Language and Culture University, Beijing, China & \bf 0.820 & 0.696$_{5}$ & 0.723$_{5}$ \\
pjetro & Warsaw University of Technology & \bf 0.790 & 0.661$_{6}$ & 0.698$_{6}$ \\
LP0606 & & \bf 0.768 & 0.637$_{8}$ & 0.679$_{8}$ \\
PP08 & & \bf 0.766 & 0.637$_{9}$ & 0.684$_{7}$ \\
AUTOHOME-ORCA \cite{SemEval2019-task8-autohome-orca} & Autohome Inc., Beijing, China and  Beijing University of Posts and Telecommunications, Beijing, China &  \bf 0.745 & 0.583$_{10}$ & 0.596$_{11}$ \\
DUTH \cite{SemEval2019-task8-duth} & Democritus University of Thrace, Xanthi, Greece & \bf 0.711 & 0.563$_{11}$ & 0.604$_{10}$ \\
cococold & & \bf 0.702 & 0.543$_{12}$ & 0.594$_{12}$ \\
nothing &  & \bf 0.702 & 0.543$_{12}$ & 0.594$_{12}$ \\
chchao & & \bf 0.630 &  0.454$_{13}$ &  0.523$_{13}$ \\
CodeForTheChange \cite{SemEval2019-task8-codeforthechange} & International Institute of Information Technology, Hyderabad, Teradata and Qubole & \bf 0.630 & 0.442$_{14}$ & 0.513$_{14}$ \\
Tuefact \cite{SemEval2019-task8-tuefact} & University of T\"ubingen, T\"ubingen, Germany & \bf 0.599 & 0.360$_{15}$ & 0.348$_{15}$ \\
Reem06 & & \bf 0.549 &  0.263$_{16}$ &  0.343$_{16}$ \\
\midrule
\it Majority Class Baseline & & \it \bf 0.450 & \it 0.009 & \it 0.333 \\
\\
\end{tabular}
\caption{\textbf{Subtask A:} Results for question classification based on the official submissions, evaluated on the test set. (Some teams did not submit system description papers, and thus we have no citations for their systems.)}
\label{table:results_subtask_a}
\end{table*}

Table~\ref{table:results_subtask_b} presents the results based on the evaluation phase on the test set for predicting answer factuality labels. This subtask was more difficult as the majority class baseline was very high due to label unbalance. No team managed to improve over that baseline, but several teams had results that were very close to it.

\begin{table*}[!ht]
\small
\centering
\begin{tabular}{lp{65mm}cccc}
\\
\bf Team ID & \bf Affiliation & \bf Accuracy & \bf F1 & \bf AvgRec & \bf MAP \\
\midrule
AUTOHOME-ORCA & Autohome Inc., Beijing, China and  Beijing University of Posts and Telecommunications, Beijing, China & \bf 0.815 & 0.511$_{2}$ & 0.512$_{2}$ & 0.155$_{7}$ \\
ColumbiaNLP & Columbia University, Department Of Computer Science and Data Science Institute & \bf 0.791 & 0.524$_{1}$ & 0.635$_{1}$ & 0.134$_{8}$ \\
DOMLIN & Deutsches Forschungszentrum f\"ur K\"unstliche Intelligenz (DFKI), Saarbrucken, Germany & \bf 0.718 & 0.402$_{3}$ & 0.445$_{3}$ & 0.267$_{3}$  \\
SolomonLab & Samsung R\&D Institute India, Bangalore & \bf 0.686 & 0.375$_{4}$ & 0.403$_{4}$ & 0.333$_{2}$ \\
CodeForTheChange & International Institute of Information Technology, Hyderabad, Teradata and Qubole & \bf 0.654 & 0.325$_{5}$ & 0.326$_{5}$ & 0.156$_{6}$ \\
BLCU\_NLP & Beijing Language and Culture University, Beijing, China & \bf 0.611 & 0.296$_{6}$ & 0.317$_{6}$ & 0.222$_{4}$ \\
LP0606 & & \bf 0.548 & 0.271$_{7}$ & 0.341$_{7}$ & 0.121$_{9}$  \\
PP08 & & \bf 0.548 & 0.271$_{7}$ & 0.341$_{7}$ & 0.121$_{9}$ \\
Tuefact & University of T\"ubingen, T\"ubingen, Germany & \bf 0.527 & 0.260$_{8}$ & 0.347$_{8}$ & 0.571$_{1}$ \\
cococold & & \bf 0.439 & 0.133$_{9}$ & 0.241$_{9}$ & 0.208$_{5}$ \\
nothing & & \bf 0.439 & 0.133$_{9}$ & 0.241$_{9}$ & 0.208$_{5}$ \\
\midrule
\it Majority Class Baseline & & \it \bf 0.830 & \it 0.285 & \it 0.333 & \it 0.156  \\
\end{tabular}
\caption{\textbf{Subtask B:} Results for answer classification based on the official submissions, evaluated on the test set.}
\label{table:results_subtask_b}
\end{table*}

\section{Discussion}

In the evaluation phase of the competition, the participants had to specify one official submission and were allowed up to two contrastive submissions. In the post-evaluation phase, they could upload an unlimited number of contrastive submissions. Below, we will only discuss the official submissions. The contrastive submissions, the ablation studies, and the experiments with different techniques are described by the participants in their respective system description papers.

The best system for Subtask A was by team Fermi (IIIT Hyderabad). They used Google's Universal Sentence representation \cite{cer2018universal}, and XGBoost \cite{chen2016xgboost}.

The best system for Subtask B was by team AUTOHOME-ORCA (Autohome Inc. and  Beijing University of Posts and Telecommunications), who used BERT \cite{devlin2018bert}.

\noindent They achieved their best results by using an ensemble, and by also using question meta-information (category and subject) in addition to the question and the answer text.
They concatenated the category, the subject and the body of the questions into the first part separated by [SEP]. The replier's username and statement were concatenated as the second part. The two parts separated by [SEP] were pushed into the BERT model for answer classification. Then, based on the sequential outputs of the BERT model, some variant methods such as average-pooling, and bi-LSTM were adopted to produce the final results. To tackle the problem with insufficient training data, they further used data augmentation based on translation with Google Translate: in particular, they performed consecutive English-Chinese and Chinese-English translation to generate more synthetic training data.

Overall, the submitted systems for the two subtasks used a number of pre-processing steps to clean the text of the question and of the answer. As shown by the DOMLIN team, the pre-processing of the data turns out to be crucial. They reported up to 5\% improvement in terms of accuracy when cleaning the unannotated forum data before fine-tuning a BERT model. Common preprocessing steps included removing or replacing the URLs, the numbers, the punctuation, the symbols, spell-checking, expansion of contractions, HTML tags, etc. DUTH also used lemmatization and stopword removal.

The submitted systems used a wide range of strategies for training their models. A sizable part of the systems used manually crafted features such as linguistic, syntactic, stylistic, and semantic features.
Moreover, the systems used task-specific information such as answer ranking and rating. ColumbiaNLP also computed an average cosine similarity of one answer with respect to the other answers in the thread for subtask B, assuming that bad answers would differ substantially from the remaining answers.

While some of the approaches used character and word $n$-gram information, the teams also used word- and sentence-level embeddings. CodeForTheChange evaluated different classification algorithms fed with Skip-Thought vectors, and ultimately found that neural networks performed best for both subtasks with either concatenation or averaging over the vectors of the available texts.

\noindent Fermi performed evaluation of different embedding models - InferSent, Concatenated Power Mean Word Embedding, Lexical Vectors, ELMo and The Universal Sentence Encoder, used in subtask A to feed an XGBoost classifier. ColumbiaNLP used ULMFiT, but performed additional unsupervised tuning of the language model on questions, answers and question-answer pairs from the Qatar Living Forum. TMLab's system used the Universal Sentence Encoder.

A common neural network architecture was LSTM, where YNU-HPCC combined LSTM with an attention mechanism. TueFact used comment chain embeddings. Other machine learning algorithms that participants tried include Random Forest, Adaboost, Perceptron, and SVM, inter alia.

While for question classification (subtask A), all the necessary information was contained in the question text and in the metadata, subtask B required additional resources. Most teams used the provided additional unannotated forum data in order to pre-train their language models or to extract more data with weak supervision (DOMLIN). Furthermore, several teams used other means for data augmentation such as SQuAD (BLCU NLP) or external Web information (SolomonLab).

\section{Conclusion}

We have described SemEval 2019 Task 8 on Fact Checking in Community Question Answering Forums. We received 17 and 11 submissions for Subtask A and B, respectively. Overall, subtask A (question classification) was easier and all submitted systems managed to improve over the majority class baseline. However, Subtask B (answer classification) proved to be much more challenging, and no team managed to improve over the majority class baseline, even though several teams came very close. For this latter subtask, using external resources and preprocessing proved to be crucial.

\section*{Acknowledgments}

This research is part of the Tanbih project,\footnote{\url{http://tanbih.qcri.org/}} which aims to limit the effect of ``fake news'', propaganda and media bias by making users aware of what they are reading. The project is developed in collaboration between the Qatar Computing Research Institute (QCRI), HBKU and the MIT Computer Science and Artificial Intelligence Laboratory (CSAIL).

\bibliography{semeval2019}
\bibliographystyle{acl_natbib}

\end{document}